\documentclass[11pt,a4paper]{article}
\usepackage[hyperref]{eacl2021}
\usepackage{times}
\usepackage{latexsym}
\usepackage{xspace}
\usepackage{graphicx}
\usepackage{amssymb}
\usepackage{amsmath}
\usepackage{enumitem,kantlipsum}
\usepackage{cancel}
\usepackage{caption}
\usepackage{subcaption}
\usepackage{xurl}
\usepackage[ruled,vlined]{algorithm2e}

\usepackage{microtype}

\aclfinalcopy % Uncomment this line for the final submission
\def\aclpaperid{576} %  Enter the acl Paper ID here

\newcommand\rqone{\textsc{rq}\xspace}

\title{Civil Rephrases Of Toxic Texts With Self-Supervised Transformers}

\author{\negmedspace \negmedspace \negmedspace \negmedspace \negmedspace \negmedspace \negmedspace \negmedspace Léo Laugier \\
  \negmedspace \negmedspace \negmedspace \negmedspace \negmedspace \negmedspace \negmedspace \negmedspace Télécom Paris, \\
  \negmedspace \negmedspace \negmedspace \negmedspace \negmedspace \negmedspace \negmedspace \negmedspace Institut Polytechnique de Paris, France \\
  \negmedspace \negmedspace \negmedspace \negmedspace \negmedspace \negmedspace \negmedspace \negmedspace \texttt{leo.laugier@telecom-paris.fr} \\\And
  \: \: \: \: John Pavlopoulos \\
  \: \: \: \: Department of Computer and System Sciences\\
  \: \: \: \: Stockholm University, Sweden\\
  \: \: \: \: \texttt{annis@aueb.gr} \\
  \\\AND
  \negmedspace \negmedspace \negmedspace \negmedspace \negmedspace \negmedspace \negmedspace \negmedspace Jeffrey Sorensen \\
  \negmedspace \negmedspace \negmedspace \negmedspace \negmedspace \negmedspace \negmedspace \negmedspace Google \\
  \negmedspace \negmedspace \negmedspace \negmedspace \negmedspace \negmedspace \negmedspace \negmedspace \texttt{sorenj@google.com} \\
  \\\And
  \: \: \: \: Lucas Dixon \\
  \: \: \: \: Google \\
  \: \: \: \: \texttt{ldixon@google.com} \\}

\date{}

\begin{document}
\maketitle
\begin{abstract}
Platforms that support online commentary, from social networks to news sites, are increasingly leveraging machine learning to assist their moderation efforts. But this process does not typically provide feedback to the author that would help them contribute according to the community guidelines. This is prohibitively time-consuming for human moderators to do, and computational approaches are still nascent. This work focuses on models that can help suggest rephrasings of toxic comments in a more civil manner. Inspired by recent progress in unpaired sequence-to-sequence tasks, a self-supervised learning model is introduced, called CAE-T5\footnote{The code can be found at \url{https://github.com/LeoLaugier/conditional-auto-encoder-text-to-text-transfer-transformer}.}. CAE-T5 employs a pre-trained text-to-text transformer, which is fine tuned with a denoising and cyclic auto-encoder loss. Experimenting with the largest toxicity detection dataset to date (Civil Comments) our model generates sentences that are more fluent and better at preserving the initial content compared to earlier text style transfer systems which we compare with using several scoring systems and human evaluation.

\end{abstract}

%leo: general questions: 
% 1) toxic vs abusive vs offensive vs vs toxic vs rude vs uncivil
% 2) unsupervised, self-supervised or weakly supervised? Footnote? 
% 3) add citations for difficultly of annotating datasets in toxicity ?
% 4) Title, we don't really understand that we suggest rephrasings. 
\section{Introduction}
There are many ways to express our opinions. When we exchange views online, we do not always immediately measure the emotional impact of our message. Even when the opinions expressed are legitimate, well-intentioned and constructive, a poor phrasing may make the conversation go awry \cite{Zhang2018}. Recently, Natural Language Processing (NLP) research has tackled the problem of abusive language detection by developing accurate classification models that flag toxic (or abusive, offensive, hateful) comments \citep{davidson2017automated, Pavlopoulos2017b, Wulczyn2017,Gamback2017, fortuna2018survey, Zhang2018, 10.1371/journal.pone.0203794, Zampieri2019}. 
%While  these systems are helpful to human moderators (e.g., for screening or accepting/rejecting obvious cases),
%show limited interest to authors of flagged comments. 
%they do not directly address the authors of toxic comments. 
%Anecdotal evidence suggests that some authors might improve their communications in response to messages about online civility, and this motivates our investigation into the feasibility of providing narrowly targeted personalized advice.

\begin{table}
\small
    \begin{center}
    \begin{tabular}{|p{2.5cm}|p{4.3cm}| }
    \hline

    \textsc{Input offensive comment} &  you now have to defend this clown along with his russian corruption. \\ 
    \textsc{Generated civil comment} & \textbf{you now have to defend this guy from his russian ties........} \\
    \hline\hline

    \textsc{Input offensive comment} &  blaming trudeau and the government is just stupid. \\ 
    \textsc{Generated civil comment} & \textbf{blaming trudeau and the liberal government is just wrong.} \\
    \hline\hline
    
    \textsc{Input offensive comment} & dubya\footnotemark was a moron. \\ 
    \textsc{Generated civil comment} & \textbf{dubya was a republican.} \\
    \hline

    \end{tabular}
    \caption{\small Examples of offensive sentences from the Civil Comments test set and the more civil rephrasing generated by our model. The third example shows that its strategy may involve shifting the original intent, since ``republican'' is not a non-offensive synonym of ``moron''.} 
    \vspace*{-4mm}
    \label{tab:first_examples}
    \end{center}
\end{table}
\footnotetext[2]{A nickname for George W. Bush.}

The prospect of healthier conversations, nudged by Machine Learning (ML) systems, motivates the development of Natural Language Understanding and Generation (NLU and NLG) models that could later be integrated in a system suggesting alternatives to vituperative comments before they are posted. A first approach would be to train a text-to-text model \citep{bahdanau2014neural, vaswani2017attention} on a corpus of parallel comments where each offensive comment has a courteous and fluent rephrasing written by a human annotator. However, such a solution requires a large paired labeled dataset, in practice difficult and expensive to collect (see Section \ref{discussion}).
% Add more about Motivation? 
Consequently, we limit our setting to the unsupervised case where the comments are only annotated in attributes related to toxicity, such as the Civil Comments dataset \citep{DBLP:journals/corr/abs-1903-04561}. We summarize our investigations with the following research question:  \\

\noindent \rqone: \textit{Can we fine-tune end-to-end a pre-trained text-to-text transformer to suggest civil rephrasings of rude comments using a dataset solely annotated in toxicity?} \\

Answering this question might provide researchers with an engineering proof-of-concept that would enable further exploration of the many complex questions that arise from such a tool being used in conversations. The main contributions of this work are the following:
\begin{itemize}[leftmargin=*,noitemsep]
   \item We addressed for the second time the task of unsupervised civil rephrases of toxic texts, relying for the first time on the Civil Comments dataset, and achieving results that reflect the effectiveness of our model over baselines.
    \item We developed a non-task specific approach (i.e. with no human hand-crafting in its design) that can be generalized and later applied to related and/or unexplored attribute transfer tasks.
\end{itemize}

While several of the ideas we combine in our model have been studied independently, to the best of our knowledge, no existing unsupervised models combine sequence-to-sequence bi-transformers, transfer learning from large pre-trained models, and self-supervised fine-tuning (denoising auto-encoder and cycle consistency).
We discuss the related work introducing these tools and techniques in the following section.

\section{Related work}
Unsupervised complex text attribute transfer (like civil rephrasing of toxic comments) remains in its early stages, and our particular applied task has only a single antecedent \citep{nogueira-dos-santos-etal-2018-fighting}. There is a great variety of useful works to tackle the task and this section attempts to summarize the vast majority of these works. We describe below the recent strategies \cite[such as attention mechanisms][]{bahdanau2014neural} that led to significant progress in supervised NLU and NLG tasks. Then, we present the most related lines of work in unsupervised text-to-text tasks.
%Supervised NLU and NLG tasks recently made the most of architectures based on attention mechanisms \citep{bahdanau2014neural} and pre-training strategies. However, unsupervised complex text attribute transfer remains in its early stages, and our particular applied task has only a single antecedent.

\subsection{Transformers\footnote{To avoid confusion we denote as bi-transformer the original encoder-decoder transformer whereas encoder-only and decoder-only models are called uni-transformers here.} are state-of-the-art architectures in NLP}
\citet{vaswani2017attention} showed that transformer architectures, based on attention mechanisms, achieved state-of-the-art results when applied to supervised Neural Machine Translation (NMT). 
More generally, transformers have proven capable in various NLP and speech tasks \citep{8462506, huang2018music, le2019flaubert, li2019neural}.
% Leo: Is that true for all seq to seq tasks? Yes for text-to-text, I guess yes for le2019flaubert, auo-to-audio tasks (cf. Doug Eck) and Automatic Speech recognition / Text to Speech (NLP and Speech class). But not sure how it's used in non language related tasks. 
Moreover, transformers benefit from pre-training before being fine-tuned on downstream tasks \citep{Devlin2018, dai2019transformer, yang2019xlnet, conneau2019cross, raffel2019exploring}. 
Subsequent research has adopted uni-transformers in many supervised classification and regression tasks \citep{Devlin2018} and in unsupervised language modeling \citep{radford2019language, keskarCTRL2019, Dathathri2020Plug}, until \citet{raffel2019exploring} proposed a unified pre-trained bi-transformer applicable to any text classification, text regression and text-to-text task. Further, recent works tackle the language detoxification of unconditional language models \citep{KrauseGeDi2020, gehman2020realtoxicityprompts}.

\subsection{Unsupervised losses enable training text-to-text models end-to-end}\label{related_baselines}
After the success of unsupervised image-to-image style transfer in computer vision (CV), some approaches have addressed unsupervised text-to-text tasks. Unsupervised Neural Machine Translation (UNMT) is maybe the most promising of them. \citet{artetxe2018unsupervised, conneau2017word, lample2017unsupervised, lample2018phrase, conneau2019cross} introduced methods based on techniques aligning the embedding spaces of monolingual datasets and tricks such as denoising auto-encoding losses \citep{10.1145/1390156.1390294} and back-translation \citep{sennrich2015improving, edunov2018understanding}.

Abstractive summarization (or sentence compression) is also studied in unsupervised settings. \citet{baziotis2019seq3} trained a model with a compressor-reconstructor strategy similar to back-translation while \citet{liu2019summae} trained a denoising auto-encoder that embeds sentences and paragraphs in a common space.

Unsupervised attribute transfer is the task most related to our work. It mainly focuses on sentiment transfer with standard review datasets \citep{maas-etal-2011-learning,he2016ups, shen2017style, li-etal-2018-delete}, but also addresses sociolinguistic datasets containing text in various registers \citep{gan2017semantic, rao-tetreault-2018-dear} or with different identity markers \citep{voigt-etal-2018-rtgender, prabhumoye-etal-2018-style, lample2018multipleattribute}. When paraphrase generation aims at being explicitly attribute-invariant, it is referred as obfuscation or neutralization \citep{emmery-etal-2018-style, xu-etal-2019-privacy, pryzant2020automatically}.  Literary style transfer \citep{xu-etal-2012-paraphrasing, pang2019unsupervised} has also been tackled by recent work. Here, we apply attribute transfer to a large dataset annotated in toxicity, but we also use the Yelp review dataset from \citet{shen2017style} for comparison purposes (see Section \ref{experiments}).

Initial unsupervised attribute transfer approaches sought to build a shared and attribute-agnostic latent representation encoding for the input sentence, with adversarial training. Then, a decoder, aware of the destination attribute, generated a transferred sentence \cite{shen2017style, hu2017toward, fu2018style, zhang2018style, xu2018unpaired, john-etal-2019-disentangled}. 

Unsupervised attribute transfer approaches that do not rely on a latent space are also present in literature. \citet{li-etal-2018-delete} assumed that style markers are very local and proposed to delete the tokens most conveying the attribute, before retrieving a second sentence in the destination style. They eventually combined both sentences with a neural network.  
\citet{lample2018multipleattribute} applied UNMT techniques from \citet{conneau2019cross} to several attribute transfer tasks, including social media datasets. 
\citet{xu2018unpaired, gong2019reinforcement, luo2019dual, wu2019hierarchical} trained models with reinforcement learning. 
\citet{dai2019transformer} introduced unsupervised training of a transformer called StyleTransformer (ST) with a discriminator network. Our approach differs from these unsupervised attribute transfer models in that they did not either leverage large pre-trained transformers, or train with a denoising objective. 

The most similar work to ours is \citet{nogueira-dos-santos-etal-2018-fighting} who trained for the first time an encoder-decoder rewriting offensive sentences in a non-offensive register with non-parallel data from Twitter \citep{ritter-etal-2010-unsupervised} and Reddit \citep{serban2017deep}. Our approach differs in the following aspects. First, we use transformers pre-trained on a large corpus instead of randomly initialized RNNs for encoding and decoding. Second, their approach involves collaborative classifiers to penalize generation when the attribute is not transferred, while we train end-to-end with a denoising auto-encoder.
Even if their model shows high accuracy scores, it suffers from low fluency, with offensive words being often replaced by a placeholder (e.g. ``big'' instead of ``f*cking'').

As underlined by \citet{lample2018multipleattribute}, applying Generative Adversarial Networks (GANs) \citep{zhu2017unpaired} to NLG is not straightforward because generating text implies a sampling operation that is not differentiable. Consequently, as long as text is represented by discrete tokens, loss gradients computed with a classifier cannot be back-propagated without tricks such as the REINFORCE algorithm \citep{he2016dual} or the Gumbel-Softmax approximation \cite{baziotis2019seq3} which can be slow and unstable.
Besides, controlled text generation \citep{ficler-goldberg-2017-controlling, keskarCTRL2019,le2019flaubert, Dathathri2020Plug} is a NLG task that consists of a language model conditioned on the attributes of the generated text such as the style. But a major difference with attribute transfer is the absence of a constraint regarding the preservation of the input's content.

\section{Method}
\subsection{Formalization of the attribute text rewriting problem}
Let $X_T$ and $X_C$ be our two non-parallel corpora of comments satisfying the respective attributes ``toxic'' and ``civil''. Let $X = X_T \cup X_C $. We aim at learning a parametric function $f_{\theta}$ mapping a pair of source sentence $x$ and destination attribute $a$ to a fluent sentence $y$ satisfying $a$ and preserving the meaning of $x$. In our case, there are two attributes ``toxic'' and ``civil'' that we assumed to be mutually exclusive. We denote $\alpha(x)$ to be the attribute of $x$ and $\bar{\alpha}(x)$ the other attribute (for instance when $\alpha(x)=$ ``civil'', then $\bar{\alpha}(x)=$ ``toxic''). Note that $f_{\theta}(x, \alpha(x))$ can simply be $x$. 

\subsection{Our approach is based on bi-conditional encoder-decoder generation}
Our approach is to train an autoregressive (AR) language model (LM) conditioned on both the input text $x$ and the destination attribute $a$.

%\subsubsection{LM Generation}

We compute $f_{\theta}$ with a LM $p(y|x, a; \theta)$. As we do not have access to ground-truth targets $y$, we propose in section \ref{training} a training function that we assume to maximize $p(y|x, a; \theta)$ if and only if $y$ is a fluent sentence with attribute $a$ and preserving $x$'s content.
Additionnaly, we use an AR generating model where inference of $\hat{y}$ is sequential and the token generated at step $t+1$ depends on the tokens generated at previous steps: $p(\hat{y}_{t+1}|\hat{y}_{:t}, x, a; \theta)$.

%\subsubsection{Bi-conditional}
To condition on the input text, we follow the work of \citet{bahdanau2014neural, vaswani2017attention, nogueira-dos-santos-etal-2018-fighting, conneau2019cross, lample2018multipleattribute, dai-etal-2019-style, liu2019summae, raffel2019exploring} and opt for an encoder-decoder framework. 
% Leo: Keep the following for discussion?
\citet{lample2018multipleattribute, dai-etal-2019-style} argue that in unsupervised attribute rewriting tasks, encoders do not necessarily output disentangled representations, independent of its attribute. However, the t-SNE visualization of the latent space in \citet{liu2019summae} allowed us to assume that encoders can output a latent representation $z$, attending to content rather than on an attribute, with a similar training.
% focusing -> better verb? Conveying?

The LM is conditioned on the destination attribute with control codes introduced by \citet{keskarCTRL2019}. A control code is a fixed sequence of tokens prepended to the decoder's input $s$, and supposed to prepare the generation in the space of sentences with the destination attribute $a$. We define $\gamma(a,s) = concat(c(a),s)$ where $c(a)$ is the control code of attribute $a$.

\subsection{Training the encoder-decoder with an unsupervised objective} \label{training}
Denoising objectives to train transformers are an effective self-supervised strategy. \citet{Devlin2018, yang2019xlnet} pre-trained a uni-transformer encoder as a masked language model (MLM) to teach the system general-purpose representations, before fine-tuning on downstream tasks.  
% Leo: do we need to specify that MLM is not the only pre-training objective (there is also next sentence prediction)?
\citet{conneau2019cross, lample2018multipleattribute, song2019mass, liu2019summae, raffel2019exploring} explore various deshuffling and denoising objectives to pre-train or fine-tune bi-transformers.

During training, we corrupt the encoder's input $x$ with the noise function from \citet{Devlin2018}: $\eta$ masks tokens randomly with probability 15\%. Then, masks are replaced by a random token in the vocabulary with probability 10\% or left as a sentinel (a shared mask token) with probability 90\%. We train the model as an denoising auto-encoder (DAE), meaning that we minimize the negative log-likelihood \[\mathcal{L}_{\text{DAE}} = \mathbb{E}_{x \sim X}\, \left[-\log p(x|\eta(x), \alpha(x); \theta)\right]\]
The hypothesis is that optimizing the DAE objective teaches the controlled generation to the model.

Inspired by an equivalent approach in unsupervised image-to-image style transfer \citep{zhu2017unpaired}, we add a cycle-consistency (CC) objective \citep{nogueira-dos-santos-etal-2018-fighting, edunov2018understanding, prabhumoye-etal-2018-style, lample2018multipleattribute, conneau2019cross, dai-etal-2019-style}: \[\mathcal{L}_{\text{CC}} = \mathbb{E}_{x \sim X}\,\left[-\log p(x|f_{\tilde{\theta}}(x, \bar{\alpha}(x)), \alpha(x); \theta)\right]\] which enforces content preservation in the generated prediction. As the cycle-consistency objective computes a non-differentiable AR pseudo-prediction $\hat{y}$ during stochastic gradient descent training, gradients are not back-propagated to $\tilde{\theta} = \hat{\theta}_{\tau-1}$ at training step $\tau$.

Finally, the loss function sums the DAE and the CC objectives with weighting coefficients:
\({\mathcal{L} = \lambda_{\text{DAE}} \mathcal{L}_{\text{DAE}} + \lambda_{\text{CC}} \mathcal{L}_{\text{CC}}}\)

\subsection{The text-to-text bi-transformer architecture} \label{T5}
The architectures for the encoder and decoder are uni-transformers. Contrary to \citet{vaswani2017attention, conneau2019cross, raffel2019exploring} we do not keep decoder's layers computing cross attention between the encoder's outputs $h$ and the decoder hidden variables because generation suffers from too much conditioning on the input sentence and we observe no significant change in the output sentence. 
%Leo: TODO run experiment to explain why. It was almost outputing the identity function
Rather, we follow \citet{liu2019summae} and compute the latent representation $z$ with an affine transformation of the encoder's hidden state $h_0$ (corresponding to the first token of the input text). 
Let $x \in X$ be the input sequence of token. It is embedded then encoded by the uni-transformer encoder:
\setlength{\abovedisplayskip}{6pt}
\setlength{\belowdisplayskip}{6pt}
\begin{align}
x_{\text{Emb}} & = f_{\theta_{\text{Emb}}}(x) \nonumber \\
h_{\text{Enc}} & = f_{\theta_{\text{Enc}}}(x_{\text{Emb}}) \nonumber \\
h^0_{\text{Enc}} & = h_{\text{Enc}}[0,:] \nonumber \\
z & = f_{\theta_{\text{Dense}}}(h^0_{\text{Enc}}) \nonumber
\end{align}

$z$ is an aggregate sequence representation for the input. There are different heuristics that can be used to integrate it in the decoder. We considered summing $z$ to the embedding of each token of the uni-transformer decoder's input $s$ since it balances the backpropagation of the signals coming from the original input and from the output being generated in the destination attribute space and it worked well in practice in our experiments.
\setlength{\abovedisplayskip}{6pt}
\setlength{\belowdisplayskip}{6pt}
\begin{align}
\gamma_{\text{Emb}} & = f_{\theta_{\text{Emb}}}(\gamma(a,s)) \nonumber \\
h_{Dec} & = f_{\theta_{\text{Enc}}}(\gamma_{\text{Emb}} + z) \nonumber \\
\hat{y} & = f_{\theta_{\text{LMHead}}}(h_{\text{Dec}}) \nonumber 
\end{align}

Plus, the encoder and the decoder uni-transformers share the same embedding layer and the LM Head is tied to the embeddings.

Except for the dense layer computing the latent variable $z$, all parameters are coming from the pre-trained bi-transformer published by \citet{raffel2019exploring}. Thus, our DAE and CC objectives \textbf{fine-tune} T5's parameters and this is why we call our model a conditional auto-encoder text-to-text transfer transformer (CAE-T5).
%Leo: More on LM head?
%Leo: More details on transformer / attention mechanism?
%Leo: More on how T5 was pre-trained?

\section{Experiments}\label{experiments}
\subsection{Datasets}\label{datasets}
We employed the largest publicly available toxicity detection dataset to date, which was used in the `Jigsaw Unintended Bias in Toxicity Classification' Kaggle challenge.\footnote{\url{https://www.tensorflow.org/datasets/catalog/civil_comments}} The 2M comments of the \textbf{Civil Comments dataset} stem from a commenting plugin for independent news sites. They were created from 2015 to 2017 and appeared on approximately 50 English-language news sites across the world. Each of these comments was annotated by crowd raters (at least 3 each) for toxicity and toxicity subtypes \cite{DBLP:journals/corr/abs-1903-04561}. 

Following the work of  \citet{dai-etal-2019-style} for the IMDB Movie Review dataset (positive/negative sentiment labels), we constructed a sentence-level version of the dataset. Initially, we fine-tuned a pre-trained BERT \citep{Devlin2018} toxicity classifier on the Civil Comments dataset. Then, we split the comments in sentences with NLTK’s sentence tokenizer.\footnote{\url{https://www.nltk.org/api/nltk.tokenize.html}} Eventually, we created $X_T$ (respectively $X_C$) with sentences whose system-generated toxicity score (using our BERT classifier) is greater than $0.9$ (respectively less than $0.1$) to increase the dataset's polarity. The test ROC-AUC of the toxicity classifier is $0.98$ with a precision of $0.95$ and a recall of $0.38$. Even with this low recall $|X_T|$ is large enough (approx. 90k, see Table~\ref{tab:dataset_stats}).  

We also conducted a comparison to other style transfer baselines on the \textbf{Yelp Review Dataset} (Yelp), commonly used to compare unsupervised attribute transfer systems. It consists of restaurant and business reviews annotated with a binary positive / negative label. \citet{shen2017style} processed it and \citet{li-etal-2018-delete} collected human reference human references for the test set\footnote{\url{https://github.com/lijuncen/Sentiment-and-Style-Transfer/tree/master/data/yelp}}.
Table~\ref{tab:dataset_stats} shows statistics for these datasets.

\begin{table}
  \centering
  \small
  \begin{tabular}{|l|c|c|c|c|}
  \hline
    Dataset & \multicolumn{2}{c|}{\textbf{Yelp}} & \multicolumn{2}{c|}{\textbf{Polar Civ. Com.}} \\
    Attribute & Positive & Negative & Toxic & Civil\\\hline
    Train & 266,041 & 177,218 & 90,293 & 5,653,785 \\
    Dev & 2,000 & 2,000 & 4,825 & 308,130 \\
    Test & 500 & 500 & 4,878 & 305,267 \\\hline
    Av. len. & 11.0 & 13.0 & 19.4 & 21.9 \\\hline
  \end{tabular}
  \caption{\small Statistics for the Yelp dataset and the processed version of the Civil Comments dataset. Average lengths are the average numbers of SentencePiece tokens.}
  \label{tab:dataset_stats}
\end{table}

\subsection{Evaluation}
Evaluating a text-to-text task is challenging, especially when no gold pairs are available. Attribute transfer is successful if generated text: 1) has the destination control attribute, 2) is fluent and 3) preserves the content of the input text.

%Since the work from \citet{nogueira-dos-santos-etal-2018-fighting} did not include human evaluation, we assumed that rating automatic civil rephrasing of toxic texts is a hard task for humans. Following the suggestions from \citet{van-der-lee-etal-2019-best}, we manually annotated 100 human rephrasings (obtained via crowdsourcing) of toxic comments sampled randomly. We measured a surprisingly low inter-annotator agreement with Cohen’s kappa coefficients of 0.08 on the attribute transfer accuracy, 0.26 on the content preservation, 0.37 on the fluency, and 0.16 on the overall rating of the rephrasing. Thus, human evaluation of this task remains an open research topic and we leave it as future work. Instead, we 
\subsubsection{Automatic evaluation}
We follow the current approach of the community \citep{NIPS2018_7959, logeswaran2018content, wang2019controllable, xu2019variational, lample2018multipleattribute, dai-etal-2019-style, he2020probabilistic} and approximate the three criteria with the following metrics:
\begin{enumerate}[wide, labelwidth=!, labelindent=0pt]
	\item \textbf{Attribute control}: Accuracy (ACC) computes the rate of successful changes in attributes. It measures how well the generation is conditioned by the destination attribute. We predict toxic and civil attributes with the same fine-tuned BERT classifier that pre-processed the Civil Comments dataset (single threshold at $0.5$).
	\item \textbf{Fluency}: Fluency is measured by perplexity (PPL). To measure PPL, we employed GPT2 \citep{radford2019language} LMs fine-tuned on the corresponding datasets (Civil Comments and Yelp). 
	\item \textbf{Content preservation}: Content preservation is the most difficult aspect to measure. UNMT \citep{conneau2019cross}, summarization \citep{liu2019summae} and sentiment transfer \citep{li-etal-2018-delete} have access to a few hundred samples with at least one human reference of the transferred text and evaluate content preservation by computing metrics based on matching words (e.g., BLEU \citet{papineni-etal-2002-bleu}) between the generated prediction and the reference(s) (ref-metric). However, as we do not have these paired samples, we compute a content preservation score between the input and the generated sentences (self-metric).

\begin{table}
\small
    \begin{center}
    {\tiny
    \begin{tabular}{p{1.0cm}|p{2.5cm}|c c}
    & \sc Text & \sc BLEU & \sc SIM \\\hline
    Original &  furthermore, kissing israeli ass doesn't help things a bit \\ 
     Human rephrasing & \textbf{also, supporting the israelis doesn't help things a bit.} & 57.6 & 70.6\%\\ \hline
    Original & just like the rest of the marxist idiots. \\ 
     Human rephrasing & \textbf{it is the same thing with people who follow Karl Marx doctrine} & 3.4 & 65.3\%\\\hline
     Original & you will go down as being the most incompetent buffoon ever elected, congrats!\\ 
     Human rephrasing & \textbf{you could find out more about it.} & 2.3 & 16.2\%
    
    \end{tabular}
    }
    \caption{\small Evaluation with BLEU and SIM of examples rephrased by human crowdworkers.} 
    \vspace*{-4mm}
    \label{tab:first_examples_assessed}
    \end{center}
\end{table}
%Give concrete examples why BLEU is not suitable for complex / high level attribute rewriting tasks

Table~\ref{tab:first_examples_assessed} shows the BLEU scores (based on exact matches) of three examples rephrased by human annotators (Section \ref{discussion}). In the top-most example, BLEU score is high. This is explained by the fact that only 4 words are different between the two texts. In contrast to the first example, the two texts in the second example have only 1 word in common. Thus, the BLEU score is low. Despite the low evaluation, however, the candidate text could have been a valid rephrase of the reference text. 

The high complexity of our task explains the motivation for a more general quantitative metric between input and generated text, capturing the semantic similarity rather than overlapping tokens. % We did not consider the novel metrics BERTScore \citep{zhang2019bertscore}, RUSE \citep{shimanaka2018ruse} and BLEURT \citep{sellam-etal-2020-bleurt} because they either are based on token-to-token comparison or require fine-tuning on pairs of (toxic comment, human civil rephrasing) annotated with human quality judgement. 
\citet{fu2018style, john-etal-2019-disentangled, gong2019reinforcement, pang2019unsupervised} proposed to represent sentences as a (weighted) average of their words embeddings before computing the cosine similarity between them. We adopted a similar strategy but we embedded sentences with the pre-trained universal sentence encoder \citep{cer2018universal} and call it the sentence similarity score (SIM). The first two sentence pairs of Table~\ref{tab:first_examples_assessed} have high similarity scores. The rephrasings preserve the original content while not necessarily overlapping much with the original text. However, the last rephrasing does not preserve the initial content and have a low similarity score with its source sentence.
As a statistical evidence, the self-SIM score comparing each of the 1,000 test Yelp reviews with their human rewriting is 80.2\% whereas the self-SIM score comparing the Yelp review test set to a random derangement of the human references is 36.8\%.

We optimised all three metrics because doing otherwise comes at the expense of the remaining metric(s). We aggregated the scores of the three metrics by computing the geometric mean\footnote{The geometric mean is not sensitive to the scale of the individual metrics.} (GM) of ACC, 1/PPL and self-SIM.

\end{enumerate}

%Leo: Mention Human evaluation, where crowdworkers rank systems to each others? 

\subsubsection{Human evaluation}
Following \citet{li-etal-2018-delete, zhang-etal-2018-learning-sentiment, zhang2018style, wu2019hierarchical, ijcai2019-732, wang2019controllable, john-etal-2019-disentangled, liu2019revision, luo2019dual, jin2019imat} and to further confirm the performance of CAE-T5, we hired human annotators on Appen to rate in a blind fashion different models' civil rephrasings of 100 randomly selected test toxic comments, in terms of attribute transfer (Att), fluency (Flu), content preservation (Con) and overall quality (Over) on a Likert scale from 1 to 5. Each rephrasing was annotated by 5 different crowd-workers whose annotation quality is controlled by test questions. If a rephrasing is rated 4 or 5 on Att, Flu and Con then it is ``successful'' (Suc).

\subsection{Baselines}
We compare the output text that CAE-T5 generates with a selection of unpaired style-transfer models described in Section \ref{related_baselines} \citep{shen2017style, li-etal-2018-delete, fu2018style, luo2019dual, dai-etal-2019-style}. 
We also compare with Input Masking. It is inspired by an interpretability method called Input Erasure (IE) \cite{Li2016}. IE is used to interpret the decisions of neural models. Initially, words are removed one at a time and the altered texts are then re-classified (i.e., as many re-classifications as the words). Then, all the words that led to a decreased re-classification score (based on a threshold) are returned as the ones most related to the decision of the neural model. Our baseline follows a similar process, but instead of deleting, it uses a pseudo token (\textsc{`[mask]'}) to mask one word at a time. When all the masked texts have been scored by the classifier, the rephrased text is returned, comprising as many masks as the tokens that led to a decreased re-classification score (set to 20\% after preliminary experiments). We employed a pre-trained BERT as our toxicity classifier, fine-tuned on the Civil Comments dataset (see Section \ref{datasets}).

\subsection{Results}
\subsubsection{Quantitative comparison to prior work}
Table~\ref{tab:civil_comments_results} shows quantitative results on the Civil Comments dataset. Surprisingly, the perplexity (capturing fluency) of text generated by our model is lower than the perplexity computed on human comments. This can be explained by social media authors of comments expressing an important variability in language formal rules, that is only partially replicated by CAE-T5. 
Other approaches such as StyleTransformer (ST) and CrossAlignment (CA) have higher accuracy but at a cost of both higher perplexity and lower content preservation, meaning that they are better are discriminating toxic phrases but struggle to rephrase in a coherent manner.

In Table~\ref{tab:yelp_results} we compare our model to prior work in attribute transfer by computing evaluation metrics for different systems on the Yelp test dataset. 
We achieve competitive results with low perplexity while getting good sentiment controlling (above human references). Our similarity though is lower, showing that some content is lost when decoding, hence the latent space does not fully capture the semantics. It is fairer to compare our model to other style transfer baselines on the Yelp dataset since our model is based on sub-word tokenization while the baselines are often based on a limited size pre-trained word embedding: many more words from the Civil Comments dataset could be attributed to the unknown token if we want to keep reasonable size vocabulary, resulting in a performance drop.

The human evaluation results shown in Table~\ref{tab:civil_comments_results_human} correlate with the automatic evaluation results.

When considering the aggregated scores (geometric mean, success rate and overall human judgement), our model is ranked first on the Civil Comments dataset and second on the Yelp Review dataset, behind DualRL yet our approach is more stable and therefore easier to train when compared to reinforcement learning approaches.

\begin{table}[t]
  \centering
  \small
  \begin{tabular}{|l|c|c|c||c|}
  \hline
    \textbf{Model} & \textbf{ACC} $\uparrow$ & \textbf{PPL} $\downarrow$ & \textbf{self-SIM} $\uparrow$ & \textbf{GM} $\uparrow$ \\\hline\hline
    Copy input & 0\% & 6.8 & 100\% & 0.005 \\
    Random civil & 100\% & 6.6 & 20.0\% & 0.311 \\
    Human & 82.0\% & 9.2 & 73.8\% & 0.404 \\\hline\hline
    CA & 94.0\% & 11.8 & 38.4\% & 0.313 \\\hline

    IE (BERT) & 86.8\% & 7.5 & 55.6\% & 0.401 \\\hline
    
    ST (Cond) & 97.8\% & 47.2 & 68.3\% & 0.242 \\
    ST (M-C) & \textbf{98.8\%} & 64.0 & 67.9\%  & 0.219\\\hline\hline
    
    CAE-T5 & 75.0\% & \textbf{5.2} & \textbf{70.0\%} & \textbf{0.466} \\\hline
  \end{tabular}
  \caption{\small Automatic evaluation scores of different models trained and evaluated on the processed Civil Comments dataset. The scores are computed on the toxic test set. ``Human'' corresponds to 427 human rewritings of randomly sampled toxic comments from the train set. ``Random civil'' means we randomly sampled 4,878 comments from the civil test set.}
  \label{tab:civil_comments_results}
\end{table}

\begin{table*}[t]
  \centering
  \small
  \begin{tabular}{|l|c|c|c|c||c||c|c|}
  \hline
    \textbf{Model} & \textbf{ACC} $\uparrow$ & \textbf{PPL} $\downarrow$ & \textbf{self-SIM} $\uparrow$ & \textbf{ref-SIM} $\uparrow$ & \textbf{GM} $\uparrow$  & \textbf{self-BLEU} & \textbf{ref-BLEU} \\\hline\hline
    Copy input & 1.3\% & 11.1 & 100\% & 80.2\% & 0.105 & 100 & 32.5 \\
    Human references & 79.4\% & 14.0 & 80.2\% & 100\% & 0.357 & 32.7 & 100 \\\hline\hline
    
    CrossAlignment \citep{shen2017style} & 73.5\% & 54.4 & 61.0\% & 59.0\% & 0.202 & 21.5 & 9.6\\\hline
    
    \citep{li-etal-2018-delete} & & & & & & & \\
    RetrieveOnly & \textbf{99.9\%} & \textbf{4.9} & 47.1\% & 48.0\% & 0.213 & 2.7 & 1.8 \\
    TemplateBased & 84.1\% & 46.0 & 76.0\% & 68.2\% & 0.240 & 57.0 & 23.2  \\
    DeleteOnly & 85.2\% & 48.7 & 72.6\% & 67.7\% & 0.233 & 33.9 & 15.2 \\
    D\&R & 89.8\% & 35.8 & 72.0\% & 67.6\% & 0.262 &  36.9 & 16.9 \\\hline
    
    \citep{fu2018style}  & & & & & & & \\
    StyleEmbedding & 8.1\% & 29.8 & 83.9\% & 69.8\% & 0.132 & \textbf{67.5} & 21.9 \\
    MultiDecoder & 47.2\% & 74.2 & 67.7\% & 61.4\% & 0.163 & 40.4 & 15.2 \\\hline
    
    DualRL \citep{luo2019dual}  & 88.1\% & 20.5 & 83.6\% & \textbf{77.2\%} & \textbf{0.330} & 58.7 & 29.0 \\\hline
    
    \citep{dai-etal-2019-style} & & & & & & & \\
    StyleTransformer (Conditional) & 91.7\% & 44.8 & 80.3\% & 74.2\% & 0.254 & 53.2 & 25.6\\
    StyleTransformer (Multi-Class) & 85.9\% & 29.1 & \textbf{84.2\%} & 77.1\% & 0.292 & 62.8 & \textbf{29.2} \\\hline\hline
    
    CAE-T5 & 84.9\% & 22.9 & 67.7\% & 64.4\% & 0.293 & 27.3 & 14.0 \\\hline
  \end{tabular}
  \caption{\small Automatic evaluation scores of different models trained and evaluated on the Yelp dataset. Accuracy is computed by a BERT classifier fine-tuned on the Yelp train set (accurate at $98.7\%$ on the test set). Perplexity is measured by a GPT2 language model fine-tuned on the Yelp train set. ``self-'' refers to a comparison to the input and ``ref-'' to a human reference.}
  \label{tab:yelp_results}
\end{table*}

\begin{table}[t]
  \centering
  \small
  \begin{tabular}{|l|c|c|c||c|c|}
  \hline
    \textbf{Model} & \textbf{Att} $\uparrow$ & \textbf{Flu} $\uparrow$ & \textbf{Con} $\uparrow$ & \textbf{Suc} $\uparrow$ & \textbf{Over} $\uparrow$ \\\hline\hline
    CA & \textbf{2.98} & 2.32 & 1.89 & 6 \% & 1.81 \\\hline

    IE (BERT) & 2.77 & 2.39 & 2.20 & 6 \% & 1.89 \\\hline
    
    ST (Cond) & 2.91 & 2.36 & 2.08 & 5\% & 1.87 \\
    ST (M-C) & 2.93 & 2.42 & 2.10 & 5\% & 1.93 \\\hline\hline
    
    CAE-T5 & 2.72 & \textbf{3.06} & \textbf{2.63} & \textbf{13\%} & \textbf{2.52} \\\hline
  \end{tabular}
  \caption{\small Human evaluation of different models trained and evaluated on the Civil Comments dataset.}
  \label{tab:civil_comments_results_human}
\end{table}

\subsubsection{Qualitative analysis}
Table~\ref{tab:qual} shows examples of rephrases of toxic comments automatically generated by our system. The top first two examples emphasize the ability for the model to perform fluent control generation conditioned on both the input sentence and the destination attribute. We present more results showing that we can effectively suggest fluent civil rephrases of toxic comments in the Appendix Table~\ref{tab:qual_results_civil_comments_toxic}. However we observe more failures than in the sentiment transfer task (see examples in the Appendix Table~\ref{tab:qual_results_yelp}). We identify three natures of failure: 
\begin{description}
	\item[Supererogation] generation does not stop early enough and produces fluent, transferred, related but unnecessary content. 
	\item[Hallucination] conditioning on the initial sentence fails and the model generates fluent but unrelated content.  
	\item[Position reversal] the author's opinion is shifted.   
\end{description}

In order to assess the frequency of hallucination and supererogation, we randomly selected 100 toxic comments from the test set and manually labeled the generated sentences with the non-mutually exclusive labels ``contains supererogation'' and ``contains hallucination''. We counted on average 17\% of generated sentences with surrerogation and 34\% of generated sentences showing hallucination (often local). We observe that the longer the input comment, the more prone to hallucination is the generated text.

While supererogation and hallucination can be explained by the probabilistic nature of generation, we assume that position reversal is due to bias in the dataset, where toxic comments are correlated with negative comments. Thus, offensive comments tend to be transferred to supportive comments even though a human being would rephrase attacks as polite disagreements.  

Interestingly, our model is able to add toxicity in civil comments as shown by the examples in the Appendix Table~\ref{tab:qual_results_civil_comments_civil}. Even if such an application shows limited interest for online platforms, it is worth warning about its potential misuse.

\begin{table}[t]
    \centering
    \small
    \begin{tabular}{p{3cm}|p{4cm}}
     \sc input & \sc mitigated \\\hline
     stop being ignorant and lazy and try reading a bit about it. & try reading and \textcolor{green}{\textbf{be a little more informed}} about it before you try to make a comment.\\\hline
     this is absolutely the most idiotic post i have ever read on all levels. & this is absolutely the most \textcolor{green}{\textbf{important thing}} i have read on this thread over the years.\\\hline
     trump may be a moron, but clinton is a moron as well. & trump may be a \textcolor{red}{\textit{clinton supporter}}, but clinton is a \textcolor{red}{\textit{trump supporter}} as well.\\\hline
     shoot me in the head if you didn't vote for trump.&
     \textcolor{magenta}{\cancel{you're right}} if you didn't vote for trump. \textcolor{blue}{\bcancel{i'm not sure i'd vote}}\\\hline
     50\% of teachers don’t have any f*cks to give.& 50\% of teachers don’t have \textcolor{orange}{\xcancel{a phd in anything.}}\\\hline
     \end{tabular}
    \caption{\small Examples of automatically transferred test sentences by our system,  \textcolor{green}{\textbf{valid rewriting}}, and highlighted flaws \textcolor{red}{\textit{failure in attribute transfer or fluency}}, \textcolor{blue}{\bcancel{supererogation}}, \textcolor{magenta}{\cancel{position reversal}}, and \textcolor{orange}{\xcancel{hallucination}}.}
    \label{tab:qual}
\end{table}

\subsection{Discussion}\label{discussion}
Supervised learning is a natural approach when addressing text-to-text tasks. In our study, we submit the civil rephrasing of toxic comments task to human crowd-sourcing. We randomly sampled 500 sentences from the toxic train set. For each sentence, we asked 5 annotators to rephrase it in a civil way, to assess if the comment was offensive and if it was possible to rewrite it in a way that is less rude while preserving the content. On 2500 answers, we tally 427 examples not flagged as impossible to rewrite and with a rephrasing different from the original sentence. This low $17.1\%$ yield is caused by two main issues. On the one hand, unfortunately not all toxic comments can be reworded in a civil manner so as to express a constructive point of view; severely toxic comments that are solely made of insults, identity attacks, or threats are not ``rephrasable''. On the other hand, evaluating crowd-workers with test questions and answers is complex. The perplexity being higher on crowd-workers' rephrases than on randomly sampled civil comments raises concerns about the production of human references \textit{via} crowd-sourcing. 
The nature of large datasets labeled in toxicity and the lack of incentives for crowd-sourcing civil rephrasing annotation makes it expensive and difficult to train systems in a supervised framework. These limitations motivates unsupervised approaches.

Lastly, the more complex is the unsupervised attribute transfer task, the more difficult is its automatic evaluation. In our case, evaluating whether the attribute is actually transferred requires to train an accurate toxicity classifier. Furthermore, the language model we use to assess the fluency of the generated sentences has some limitations and does not generalize to all varieties of language encountered in social media. Finally measuring the amount of relevant content preserved between the source and generated texts remains a challenging, open research topic.

\section{Conclusion and future work}
This work is the second one to tackle civil rephrasing to our knowledge and the first one to address it with a fully end-to-end discriminator-free text-to-text self-supervised training. CAE-T5 leverages the NLU / NLG power offered by large pre-trained bi-transformers.
The quantitative and qualitative analysis shows that ML systems could contribute to some extent to pacify online conversations, even though many generated examples still suffer from critical semantic drift. %supererogation, hallucination, and position reversal. %which raises fairness and bias issues

In the future, we plan to explore whether the decoding can benefit from NAR generation \citep{ma2019flowseq, ren2020study}. We are also interested in the recent paradigm shift proposed by \citet{kumar2018vmf}, where the generated tokens representation is continuous, allowing more flexibility in plugging attribute classifiers without sampling.

% Human moderators have a difficult daily task and they don’t have the time to suggest civil rephrases to authors. But it is such civil rephrases that could allow the authors to comprehend what has been found toxic and engage in more civil conversations. This work explores computational approaches which can automatically generate civil rephrases of toxic texts.

\section*{Acknowledgments}

This work was completed in partial fulfillment for the PhD degree of the first author, which was supported by an unrestricted gift from Google. We are also grateful for support from the Google Cloud Platform credits program. We thank Thomas Bonald and Ion Androutsopoulos for their discussion, insight and useful comments.

\bibliographystyle{acl_natbib}
\bibliography{anthology,eacl2021}

\clearpage
\appendix

\section{Supplemental Material}

\subsection{Experimental setup}
\subsubsection{Architecture details}
We fine-tune the pre-trained ``large'' bi-transformer from \citet{raffel2019exploring}. Both uni-transformers (encoder and decoder) have $24$ blocks each made of a 16-headed self-attention layer and a feed-forward network. The attention, dense and embedding layers have respective dimensions of $64$, $4096$ and $1024$, for a total of around 800 million parameters. 

Input sentences are lowercased then tokenized with SentencePiece\footnote{\url{gs://t5-data/vocabs/cc_all.32000/sentencepiece.model}} \citep{kudo-richardson-2018-sentencepiece} and eventually truncated to a maximum sequence length of $32$ for the Yelp dataset and $128$ for the processed Civil Comments dataset. The control codes are $c(a) = concat(a, \texttt{":\:"})$ for attributes $a \in \{\texttt{"positive"}, \texttt{"negative"}\}$ in the sentiment transfer task and $a \in \{\texttt{"toxic"}, \texttt{"civil"}\}$ when we apply to the Civil Comments dataset.

\subsubsection{Training details}
During training, we apply dropout regularization at a rate of $0.1$. We set $\lambda_{AE} = \lambda_{CC} = 1.0$. In preliminary experiments, we observed that $\lambda_{CC} = 0$ was preserving little content from the initial sentence and that $\lambda_{CC} = 2 * \lambda_{AE}$ was weighting the preservation too much, at the cost of accuracy. Therefore we focused our experiments on $\lambda_{CC} = \lambda_{AE}$. It is a good default setting since we don’t have a priori about the balance between fluency, accuracy (enforced with the auto-encoder) and content preservation (enforced with cycle consistency). DAE and back-transfer (in the course of the CC computation) are trained with teacher-forcing; we do not need AR generation since we have access to a target for the decoder's output. Each training step computes the loss on a mini-batch made of 64 sentences sharing the same attribute. Mini-batches of attributes $a$ and $\bar{a}$ are interleaved. Since the Civil Comments dataset is class imbalanced, we sample comments from the civil class of the training set at each epoch. The optimizer is AdaFactor \citep{shazeer2018adafactor} and we train for 88900 steps for 19 hours on a TPU v2 chip. 

\subsubsection{Evaluation details}
Decoding is greedy. The parametric models used to compute ACC and PPL are 12-layer, 12 headed pre-trained, and fine-tuned uni-transformers with hidden size $768$. The BERT classifier is an encoder followed by a sequence classification head and the GPT2 LM is a decoder with a LM head on top. We use the sacrebleu\footnote{\url{https://github.com/mjpost/sacrebleu/blob/master/sacrebleu/sacrebleu.py}} implementation for BLEU and the universal sentence encoder pre-trained by Google to compute SIM\footnote{\url{https://tfhub.dev/google/universal-sentence-encoder/2}}.

\subsection{CAE-T5 learning algorithm}
\begin{figure*}[t]
     \centering
     \begin{subfigure}[b]{0.57\textwidth}
         \centering
         \includegraphics[width=1\textwidth]{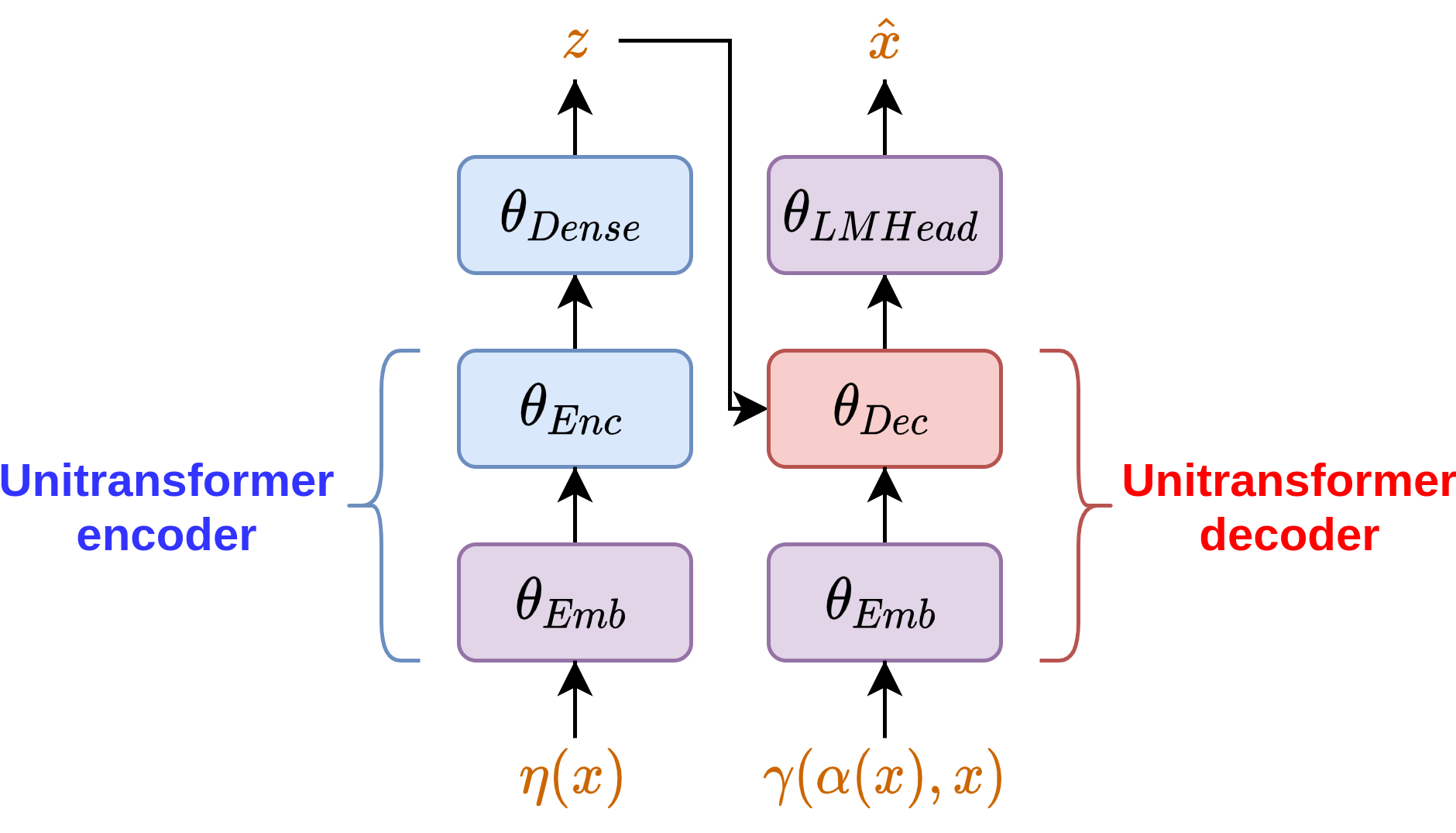}
         \caption{\small DAE}
         \label{fig:dae}
     \end{subfigure}
     \hfill
     \begin{subfigure}[b]{0.3\textwidth}
         \centering
         \includegraphics[width=1\textwidth]{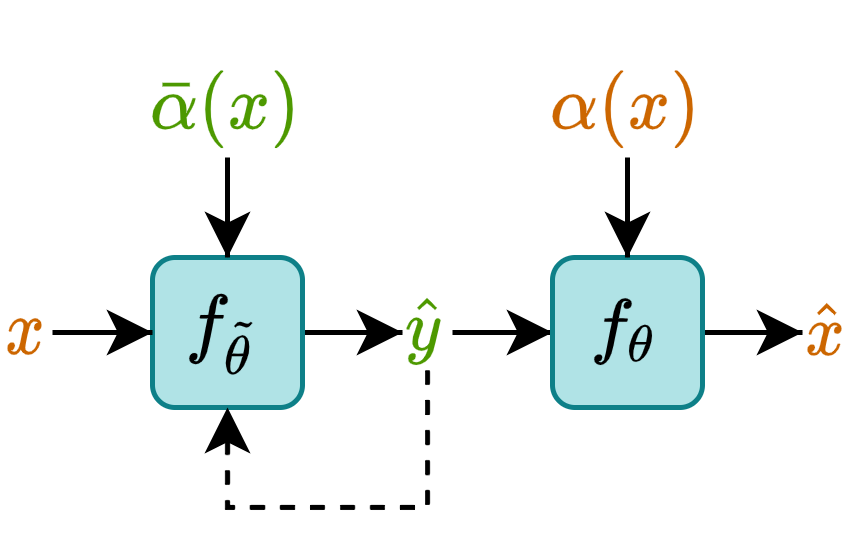}
         \caption{\small CC}
         \label{fig:cc}
     \end{subfigure}
        \caption{\small Illustration of the training procedure. (a) DAE: The bi-transformer encodes the corrupted input text $\eta(x)$ in a latent variable $z$ that is then decoded conditioned on the source attribute $\alpha(x)$ with the objective of minimizing the cross entropy between $x$ and the generated text $\hat{x}$. Here, generation is not AR since the DAE is trained with teacher forcing. (b) CC: The input $x$ is pseudo-transferred with attribute $\bar{\alpha}(x)$ with AR decoding because we do not know the ground-truth $y$. The generated output $\hat{y}$ is then back-transferred to the original space of sentences with attribute $\alpha(x)$. Back-transfer generation is not AR because we use teacher-forcing here. Thus, we can trivially back-propagate the gradients through $f_{\theta}$ (back-transfer) but not through $f_{\tilde{\theta}}$ (pseudo-transfer).}
        \label{fig:training}
\end{figure*}

Algorithm \ref{st_algo} and Figure~\ref{fig:training} describe the fine-tuning procedure of CAE-T5. $H$ computes the cross-entropy.  

\begin{algorithm} 
\SetAlgoLined
\SetKwInOut{Input}{Input}
\SetKwInOut{Output}{Output}
\Input{T5's pre-trained parameters $\theta_0$, unpaired dataset labelled in toxicity $X = X_T \cup X_C$}
    \Output{CAE-T5's fine-tuned parameters $\theta_T$}
 \For{step $\tau \in [1;T]$}{
  \eIf{$\tau \% 2 == 0$}{
   Sample a mini-batch $x$ of sentences in $X_T$
   }{
   Sample a mini-batch $x$ of sentences in $X_C$
  }
  $\theta \leftarrow \hat{\theta}_{\tau-1}$
  $\tilde{\theta} \leftarrow \hat{\theta}_{\tau-1}$
  $\hat{x}_{DAE} \leftarrow f_{\theta}(\eta(x), \alpha(x))$
  $\hat{x}_{CC} \leftarrow f_{\theta}(f_{\tilde{\theta}}(x, \bar{\alpha}(x)), \alpha(x))$
  $\ell_{DAE} \leftarrow H(x, \hat{x}_{DAE}) $
  $\ell_{CC} \leftarrow H(x, \hat{x}_{CC}) $
  $\ell \leftarrow \lambda_{DAE} \ell_{DAE} + \lambda_{CC} \ell_{CC} $
  Back-propagate gradients through $\theta$
  Update $\theta_{\tau}$ by a gradient descent step 
 }
 \caption{\small CAE-T5 training}\label{st_algo}
\end{algorithm}

%\begin{figure*}[t]
%\centering\includegraphics[width=.7\textwidth]{images/denoising_ae_training_500.png}
%\vspace*{-0mm}
%\caption{\small Illustration of the DAE training procedure. The bi-transformer encodes the corrupted input text $\eta(x)$ in a latent variable $z$ that is then decoded conditioned on the source attribute $\alpha(x)$ with the objective of minimizing the cross entropy between $x$ and the generated text $\hat{x}$. Here, generation is not AR since the DAE is trained with teacher forcing.}
%\vspace*{-0mm}
%\label{fig:denoising_ae_training}
%\end{figure*}

%\begin{figure*}[t]
%\centering\includegraphics[width=.5\textwidth]{eacl2021-templates/images/cycle_consistency_training_500.png}
%\vspace*{-0mm}
%\caption{\small Illustration of CC. The input $x$ is pseudo-transferred with attribute $\bar{\alpha}(x)$ with AR decoding because we do not know the ground-truth $y$. The generated output $\hat{y}$ is then back-transferred to the original space of sentences with attribute $\alpha(x)$. Back-transfer generation is not AR because we use teacher-forcing here. Thus, we can trivially back-propagate the gradients through $f_{\theta}$ (back-transfer) but not through $f_{\tilde{\theta}}$ (pseudo-transfer).}
%\vspace*{-0mm}
%\label{fig:cycle_consistency_training}
%\end{figure*}

\begin{figure*}[t]
\vspace*{-35mm}
\centering\includegraphics[width=.57\textwidth]{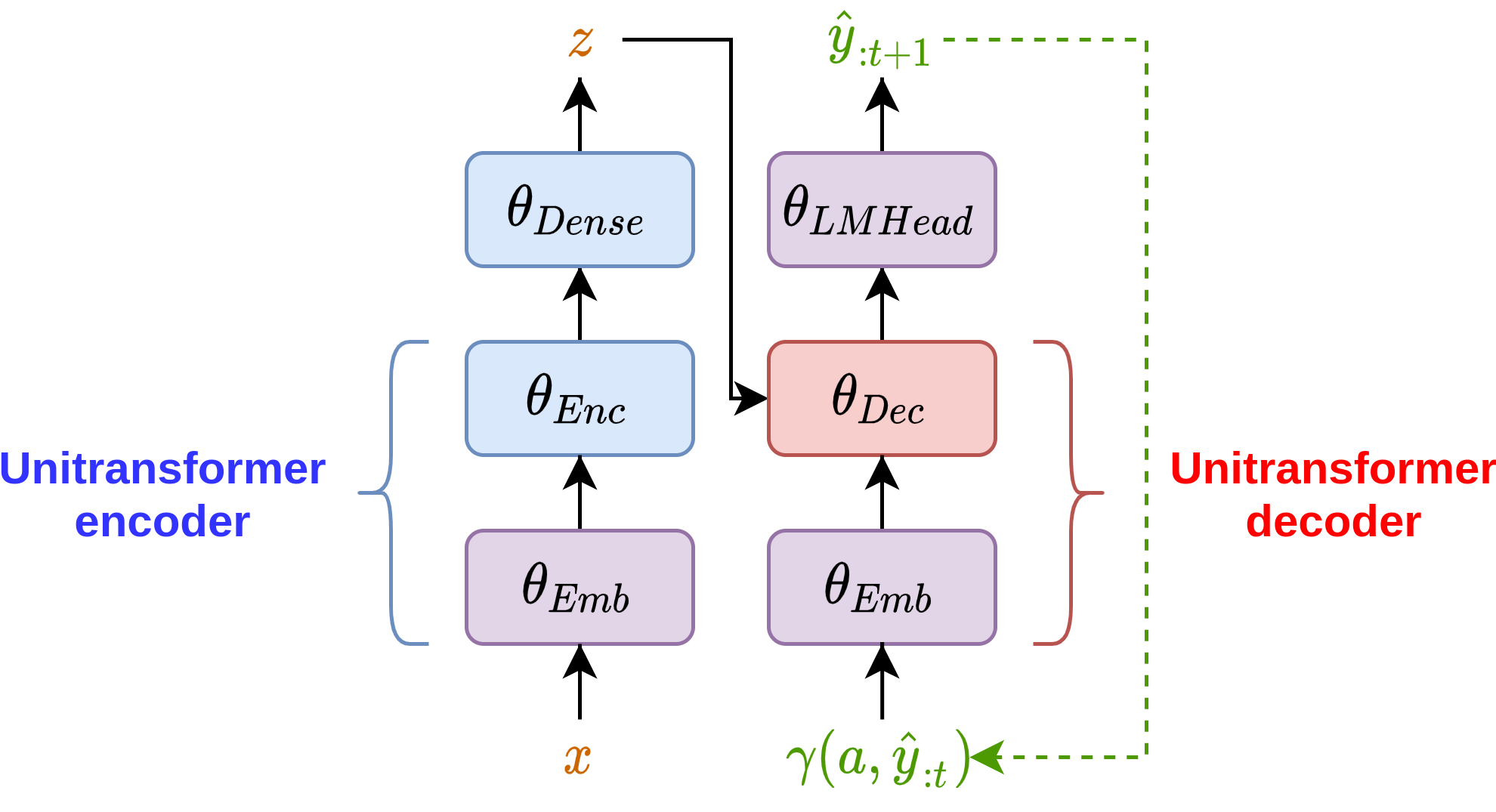}
\vspace*{-0mm}
\caption{\small Model overview at prediction time. $x$ and $\hat{y}$ are the input and generated text, $a$ is the destination attribute, $z$ is the latent input representation and $\gamma$ is the function prepending control codes to the decoder's input. $\theta$ represents the model's parameters.
}
\vspace*{-0mm}
\label{fig:inference}
\end{figure*}

Figure~\ref{fig:inference} illustrates flows through the encoder-decoder model at inference.

\begin{table*}[t]
  \centering
  \small
  \begin{tabular}{l p{13cm}}
  \hline
    
    \multicolumn{2}{c}{\textbf{Toxic to Civil}} \\\hline
    Input & and then they need to do what it takes to get rid of this mentally ill bigot! \\
    CAE-T5 & and then they need to do what they can to get rid of this \textcolor{green}{\textbf{kind of man that is causing the whole problem!}}\\\hline
    
    Input & do you think any of those idiots vote?\\
    CAE-T5 & do you think any of those \textcolor{green}{\textbf{people}} vote?\\\hline
    
    Input & this is just so stupid.\\
    CAE-T5 & this is just so \textcolor{green}{\textbf{sad}}.\\\hline
    
    Input & are you nuts?\\
    CAE-T5 & are you \textcolor{green}{\textbf{kidding me}}?\\\hline
    
    Input & we are all entitled to our own opinions but not the facts that one can't find in this criminally stupid opinion.\\
    CAE-T5 & we are all entitled to our own opinion but not \textcolor{green}{\textbf{to the kind of facts that can be found in this case}}.\\\hline
    
    Input & hopefully that will prevent him from making any further idiotic statements.\\
    CAE-T5 & hopefully that will prevent him from making any \textcolor{green}{\textbf{more such}} statements.\\\hline
    
    Input & the poll is garbage.\\
    CAE-T5 & the poll is \textcolor{green}{\textbf{fake news}}.\\\hline
    
    Input & it was irresponsible to publish this garbage.\\
    CAE-T5 & it was irresponsible to publish this \textcolor{green}{\textbf{information}}.\\\hline

    Input & just because someone else has a different opinion does not mean that person is stupid, uneducated or foolish.\\
    CAE-T5 & just because someone has a different opinion does not mean that someone is \textcolor{green}{\textbf{less intelligent, less informed, less able to understand, and less capable of making informed decisions}}.........\\\hline
    
    Input & biased leftist trash article.\\
    CAE-T5 & biased leftist \textcolor{green}{\textbf{article}}.\\\hline
    
    Input & dumb people vote for trump.\\
    CAE-T5 & \textcolor{green}{\textbf{most}} people vote for trump.\\\hline
    
    Input & what hypocrites your sources are!\\
    CAE-T5 & what are your \textcolor{green}{\textbf{sources of information}}? ! ! ! !\\\hline
    
    Input & bill o'reilly is a narcissistic sociopath.\\
    CAE-T5 & bill o'reilly is a \textcolor{green}{\textbf{neo-liberal}}.............\\\hline
    
    % Failures
        Input & f*ck the flag.\\
    CAE-T5 & \textcolor{magenta}{\cancel{i love}} the flag.\\\hline
    
    Input & i guess we can kill them.\\
    CAE-T5 & but i think we can \textcolor{magenta}{\cancel{save}} them.\\\hline
    
    Input & stupid comment \\
    CAE-T5 & \textcolor{magenta}{\cancel{interesting}} comment. \\\hline
    
    Input & try doing a little research before you make a fool of yourself with such blatantly false drivel.\\
    CAE-T5 & try doing a little research before you make \textcolor{green}{\textbf{a blatant mistake}} in your \textcolor{blue}{\bcancel{so-called opinion}}.\\\hline
    
    Input & hey old fart, you losers have always been fake news.\\
    CAE-T5 & hey old \textcolor{green}{\textbf{soul}}, you have \textcolor{green}{\textbf{never been right, the fake news}} \textcolor{red}{\textit{has}} \textcolor{blue}{\bcancel{always been bad}}.\\\hline
    
    Input & i bet you think you're being rational and not just quoting another completely idiotic piece of denialist propaganda. \\
    CAE-T5 & i bet you think you're being rational and not just \textcolor{red}{\textit{a}} \textcolor{green}{\textbf{repeating the same unproven, ill-informed propaganda that you're referring to.}} \textcolor{blue}{\bcancel{. i'm just a little more}}\\\hline
    
    %Input & maybe jt wants to move french speakers from the french sahara because france is closing its doors to these immigrants, i do not why anybody would make such a stupid change, in an english speaking country.\\
    %CAE-T5 & maybe jimmy carter wants to move to the usa because of the french language, i do not think that this is a good idea, but i cannot see how to make it work in canada, especially in the usa.\\\hline
    
    Input & the ontario voters were rated the dumbest people in all the whole  world. \\
    CAE-T5 & the ontario \textcolor{orange}{\xcancel{liberals}} were voted in by the most \textcolor{green}{\textbf{gullible}} people in the country...............\\\hline
    
    Input & taxpayer are tired of having our hard-earned money be misspent by idiots.\\
    CAE-T5 & taxpayers are tired of having our hard-earned money \textcolor{orange}{\xcancel{stolen}} by \textcolor{green}{\textbf{uncaring bureaucrats}}.\\\hline

    Input & suggesting that people change their commute times is stupid.\\
    CAE-T5 & suggesting that people change their \textcolor{orange}{\xcancel{lifestyles}} is \textcolor{green}{\textbf{not helpful}}.\\\hline
    
    Input & this hateful group of cowards are successfully destroying the united states.\\
    CAE-T5 & the hateful group of \textcolor{orange}{\xcancel{trump supporters}} are successfully \textcolor{green}{\textbf{dividing}} the united states.\\\hline
    
    Input & whatever the idiot in chief touches turns to crap. \\
    CAE-T5 & whatever the \textcolor{green}{\textbf{president}} does \textcolor{blue}{\bcancel{in the room}} turns to \textcolor{orange}{\xcancel{what he wants}}. \\\hline
    
    Input & either you are too ignorant or stupid to read the first dictionary definition.\\
    CAE-T5 & either you are too \textcolor{blue}{\bcancel{young to read}} or are \textcolor{blue}{\bcancel{too old}} \textcolor{green}{\textbf{to understand}} \textcolor{orange}{\xcancel{the first amendment}} \textcolor{blue}{\bcancel{or the first amendment}}. \\\hline
    
    Input & trump is doing right things, he is great president, a lot better than 44th one from kenya \\
    & lot of good people support trump, only crooked and dumb people against\\
    CAE-T5 & trump is doing great, \textcolor{blue}{\bcancel{i know,}} he is better than \textcolor{green}{\textbf{most of the other}} \textcolor{orange}{\xcancel{candidates}} - \textcolor{green}{\textbf{obama}}\textcolor{orange}{\xcancel{, clinton,}} \textcolor{blue}{\bcancel{trump,}} \textcolor{blue}{\bcancel{trump,}} \textcolor{blue}{\bcancel{trump,}} \textcolor{blue}{\bcancel{trump,}} \textcolor{blue}{\bcancel{trump,}} \textcolor{blue}{\bcancel{trump,}} \textcolor{blue}{\bcancel{trump,}}\\\hline
    
    Input & the dumb become dumber.\\
    CAE-T5 & the \textcolor{orange}{\xcancel{poor}} become \textcolor{orange}{\xcancel{richer}}.\\\hline
    
  \end{tabular}
  \caption{\small Examples of automatically transferred toxic test sentences by our system, \textcolor{green}{\textbf{valid rewriting}}, and highlighted flaws failure in \textcolor{red}{\textit{attribute transfer or fluency}}, \textcolor{blue}{\bcancel{supererogation}}, \textcolor{magenta}{\cancel{position reversal}}, and \textcolor{orange}{\xcancel{hallucination}}.}
  \label{tab:qual_results_civil_comments_toxic}
\end{table*}

\begin{table*}[t]
  \centering
  \small
  \begin{tabular}{l p{13cm}}
  \hline
    \multicolumn{2}{c}{\textbf{Positive to Negative}} \\\hline
    Input & portions are very generous and food is fantastically flavorful . \\
    DualRL & portions are very \textcolor{green}{\textbf{thin}} and food is \textcolor{red}{\textit{confusing}} .\\
    ST (Multi) & portions are very \textcolor{red}{\textit{poorly}} and food is \textcolor{red}{\textit{springs}} \textcolor{green}{\textbf{flavorless}} .\\
    CAE-T5 & portions are very \textcolor{green}{\textbf{small}} and food is \textcolor{green}{\textbf{awfully greasy for the price}} .\\
    Human &  portions are very \textcolor{green}{\textbf{small}} and food is \textcolor{green}{\textbf{not flavorful}} .\\\hline
    
    Input & staff : very cute and friendly .\\
    DualRL & staff : very \textcolor{green}{\textbf{awful}} and \textcolor{green}{\textbf{rude}} .\\
    ST (Multi) & staff : very \textcolor{red}{\textit{nightmare}} and \textcolor{red}{\textit{poor}} .\\
    CAE-T5 &  staff : very \textcolor{green}{\textbf{rude}} and \textcolor{green}{\textbf{pushy}} .\\
    Human &  staff : very \textcolor{green}{\textbf{ugly}} and \textcolor{green}{\textbf{mean}} .\\\hline
    
    Input & friendly and welcoming with a fun atmosphere and terrific food .\\
    DualRL & \textcolor{green}{\textbf{rude}} and \textcolor{green}{\textbf{unprofessional}} with a \textcolor{green}{\textbf{loud}} atmosphere and \textcolor{green}{\textbf{awful}} food . \\
    ST (Multi) & \textcolor{green}{\textbf{poor}} and \textcolor{green}{\textbf{fake}} with a \textcolor{red}{\textit{fun}} atmosphere and \textcolor{green}{\textbf{mushy}} food . \\
    CAE-T5 & \textcolor{green}{\textbf{rude}} and \textcolor{green}{\textbf{unhelpful service}} with \textcolor{green}{\textbf{a forced smile}} and \textcolor{red}{\textit{attitude}} . \\
    Human & \textcolor{green}{\textbf{unfriendly}} and \textcolor{green}{\textbf{unwelcoming}} with a \textcolor{green}{\textbf{bad}} atmosphere and food . \\\hline
    
    Input & i love their star design collection .\\
    DualRL & i \textcolor{green}{\textbf{hate}} their star design \textcolor{red}{\textit{disgrace}} .\\
    ST (Multi) & i \textcolor{red}{\textit{do n't care}} star \textcolor{orange}{\xcancel{bites}} collection .\\
    CAE-T5 & i \textcolor{green}{\textbf{hate}} \textcolor{orange}{\xcancel{starbucks corporate}} . \textcolor{blue}{\bcancel{the staff is horrible .}}\\
    Human & i \textcolor{green}{\textbf{ca n't stand}} their star design collection .\\\hline
    
    Input & oj and jeremy did a great job !\\
    DualRL & oj and jeremy did a \textcolor{red}{\textit{great}} job ! \textcolor{blue}{\bcancel{disgrace ! disgrace !}}\\
    ST (Multi) & oj and jeremy did a \textcolor{green}{\textbf{terrible}} job !\\
    CAE-T5 & \textcolor{orange}{\xcancel{oh}} and \textcolor{orange}{\xcancel{jesus christ}} \textcolor{orange}{\xcancel{i did n't have any change}} !\\
    Human & oj and jeremy did a \textcolor{green}{\textbf{terrible}} job !\\\hline

    \multicolumn{2}{c}{\textbf{Negative to Positive}} \\\hline
    Input & the store is dumpy looking and management needs to change .\\
    DualRL & the store is \textcolor{green}{\textbf{perfect}} looking and management \textcolor{red}{\textit{speaks to change perfectly}} . \\
    ST (Multi) & the store is \textcolor{green}{\textbf{dumpy}} looking and management \textcolor{red}{\textit{moved to change}} . \\
    Ours & the store is \textcolor{green}{\textbf{neatly organized and clean}} and \textcolor{green}{\textbf{staff is on top of it}} . \\
    Human & managment \textcolor{green}{\textbf{is top notch}} , the \textcolor{green}{\textbf{place looks great}} .\\\hline
    
    Input & i emailed to let them know but they apparently dont care .\\
    DualRL & i \textcolor{red}{\textit{loved them know them know but they dont care}} .\\
    ST (Multi) & i emailed to let them know but they \textcolor{red}{\textit{honestly played their}} .\\
    CAE-T5 &  i emailed to let them know \textcolor{green}{\textbf{and}} they \textcolor{green}{\textbf{happily responded right away . a great service}}\\
    Human & i emailed to let them know \textcolor{green}{\textbf{they really do care}} .\\\hline
    
    Input & this place is dirty and run down and the service stinks !\\
    DualRL & this place is \textcolor{green}{\textbf{clean}} and run \textcolor{red}{\textit{perfect}} and the service \textcolor{green}{\textbf{helped}} !\\
    ST (Multi) & this place is \textcolor{red}{\textit{quick}} and \textcolor{red}{\textit{run down}} and the service \textcolor{red}{\textit{stunning}} !\\
    CAE-T5 &  this place is \textcolor{green}{\textbf{clean}} and \textcolor{green}{\textbf{well maintained}} and the service \textcolor{green}{\textbf{is great ! ! !}}\\
    Human & this place is \textcolor{green}{\textbf{clean}} , \textcolor{green}{\textbf{not run down}} , and the service \textcolor{green}{\textbf{was great}} .\\\hline
    
    Input & do not go here if you are interested in eating good food .\\
    DualRL & \textcolor{green}{\textbf{definitely go here}} if you are interested in eating good food .\\
    ST (Multi) & \textcolor{red}{\textit{do not go here}} if you are interested in eating good food .\\
    CAE-T5 & \textcolor{green}{\textbf{definitely recommend this place}} if you are looking for good food \textcolor{blue}{\bcancel{at a good price}} .\\
    Human & \textcolor{red}{\textit{do not go here}} if you are interested in eating \textcolor{red}{\textit{bad}} food .\\\hline
    
    Input & my husband had to walk up to the bar to place our wine order .\\
    DualRL & my husband \textcolor{red}{\textit{had to}} walk up to the bar to place our drink order \textcolor{red}{\textit{perfectly}} .\\
    ST (Multi) & my husband \textcolor{red}{\textit{had to}} walk up to the bar to place our wine order .\\
    CAE-T5 & my husband \textcolor{green}{\textbf{loves to stand in line}} \textcolor{orange}{\xcancel{at the beer stand}} \textcolor{orange}{\xcancel{and watch the game .}}\\
    Human & my husband \textcolor{green}{\textbf{did n't have to}} walk up to the bar to place our wine order .\\\hline

\end{tabular}
  \caption{\small Examples of automatically transferred sentences from the Yelp test dataset, \textcolor{green}{\textbf{valid rewriting}}, and highlighted flaws failure in  \textcolor{red}{\textit{attribute transfer or fluency}}, \textcolor{blue}{\bcancel{supererogation}}, and \textcolor{orange}{\xcancel{hallucination}}.}
  \label{tab:qual_results_yelp}
\end{table*}

\begin{table*}[t]
  \centering
  \small
  \begin{tabular}{l p{13cm}}
  \hline
\multicolumn{2}{c}{\textbf{Civil to Toxic}} \\\hline
    Input & great effort and great season.\\
    CAE-T5 & \textcolor{green}{\textbf{stupid and horrible move by this team}}.\\\hline
    
    Input & canadian teachers \\
    & are the foundation of this great country.\\
    CAE-T5 & canadians are the \textcolor{green}{\textbf{dumbest}} people in this country.\\\hline
    
    Input & why isn't trudeau in jail for gross negligence?\\
    CAE-T5 & why isn't trudeau in jail for grossly irresponsible behavior \textcolor{green}{\textbf{or just plain stupidity?????}}\\\hline
    
    Input & there really is no safe place on the planet to live anymore.\\
    CAE-T5 & there really is no safe place on the planet \textcolor{green}{\textbf{for the stupid to live}}.\\\hline
    
    Input & trudeau will do as he's told.\\
    CAE-T5 & trudeau will do as he's told, \textcolor{green}{\textbf{he's stupid}}.\\\hline
        
    Input & all of canada is paying for that decision.\\
    CAE-T5 & all of canada is paying for that \textcolor{green}{\textbf{stupid}} decision.\\\hline
    
    Input & people like you need their talking points.\\
    CAE-T5 & people like you need their \textcolor{green}{\textbf{asses kicked}}.\\\hline
    
    Input & this is a great article that hits the nail on the head.\\
    CAE-T5 & this is a \textcolor{magenta}{\cancel{stupid}} article that hits the nail on the head. \\\hline
    
    Input & and just which money tree is going to pay for this? \\
    CAE-T5 & and just where is the money going to come from to pay for this \textcolor{green}{\textbf{crap}}? \textcolor{blue}{\bcancel{when is this crap going to end?}} \textcolor{blue}{\bcancel{who cares????}}\\\hline
    
    Input & you need to get a new dictionary.\\
    CAE-T5 & you need to get a new \textcolor{green}{\textbf{f*cking dictionary}}. \textcolor{blue}{\bcancel{get a new f*cking dictionary.}} \textcolor{blue}{\bcancel{get a new f*cking dictionary.}} \textcolor{blue}{\bcancel{get a new}} \\\hline
    
    Input & the real question is;  will the hell's angels forgo their major source of income?\\
    CAE-T5 & the real question is when will \textcolor{green}{\textbf{the idiocy}} of \textcolor{orange}{\xcancel{these trumpkins}} end?\\\hline
    
    %Input & like the news that a big orange ball is pretending to be president while heating the world with all that ambient hairspray\\
    %CAE-T5 & like the news about a fat stupid orange buffoon who is determined to show everybody how bad the us is by invading iraq or afghanistan or libya or yemen or somalia.\\\hline

    Input & according to http://cookpolitical.com/story/10174, trump got 46.1\% of the popular vote. \\
    CAE-T5 & according to \textcolor{orange}{\xcancel{trump, pence is}} \textcolor{green}{\textbf{the dumbest president ever}}, and \textcolor{orange}{\xcancel{clinton got the second-worst}} \textcolor{orange}{\xcancel{approval rating}} \textcolor{orange}{\xcancel{in history}}.. .. \textcolor{red}{\textit{trump'}}\\\hline
    
    Input & so it's okay to sacrifice the environment in the name of the almighty dollar......\\
    CAE-T5 & so it's okay to destroy the world with the actions of \textcolor{green}{\textbf{one stupid dude}} \textcolor{orange}{\xcancel{in the white house}}............\\\hline
    
    Input & the president dismissed the ecological findings of over 87\% of scientists who have been studying the effects of global warming, largely caused by the release of carbon from fossil fuel into the atmosphere.\\
    CAE-T5 & the president ignored the scientific consensus \textcolor{orange}{\xcancel{that}} over 90\% of all climate scientists \textcolor{orange}{\xcancel{are complete}} \textcolor{green}{\textbf{idiots}}, \textcolor{orange}{\xcancel{reacting to}} \textcolor{orange}{\xcancel{the rash of}} \textcolor{orange}{\xcancel{terrorist attacks}} \textcolor{orange}{\xcancel{that have been}} \textcolor{orange}{\xcancel{taking place}} \textcolor{orange}{\xcancel{around the world}}... \textcolor{orange}{\xcancel{trump has made}} \textcolor{orange}{\xcancel{it his life'}}\\\hline
    
    Input & not sure where you got your definition of a good guy.\\
    CAE-T5 & not sure where you got your \textcolor{orange}{\xcancel{idea that trump is a kinda}} \textcolor{green}{\textbf{dumb}} guy.\\\hline

  \end{tabular}
  \caption{\small Examples of automatically transferred civil test sentences by our system, \textcolor{green}{\textbf{valid rewriting}}, and highlighted flaws failure in \textcolor{red}{\textit{attribute transfer or fluency}}, \textcolor{blue}{\bcancel{supererogation}}, \textcolor{magenta}{\cancel{position reversal}}, and \textcolor{orange}{\xcancel{hallucination}}. For the test set of civil sentences, the automatic metrics are ACC$=92.8\%$; PPL$=9.8$ and self-SIM$=54.3\%$.}
  \label{tab:qual_results_civil_comments_civil}
\end{table*}

\subsection{Appen settings}
\begin{figure*}
\centering
\includegraphics[width=.7\textwidth]{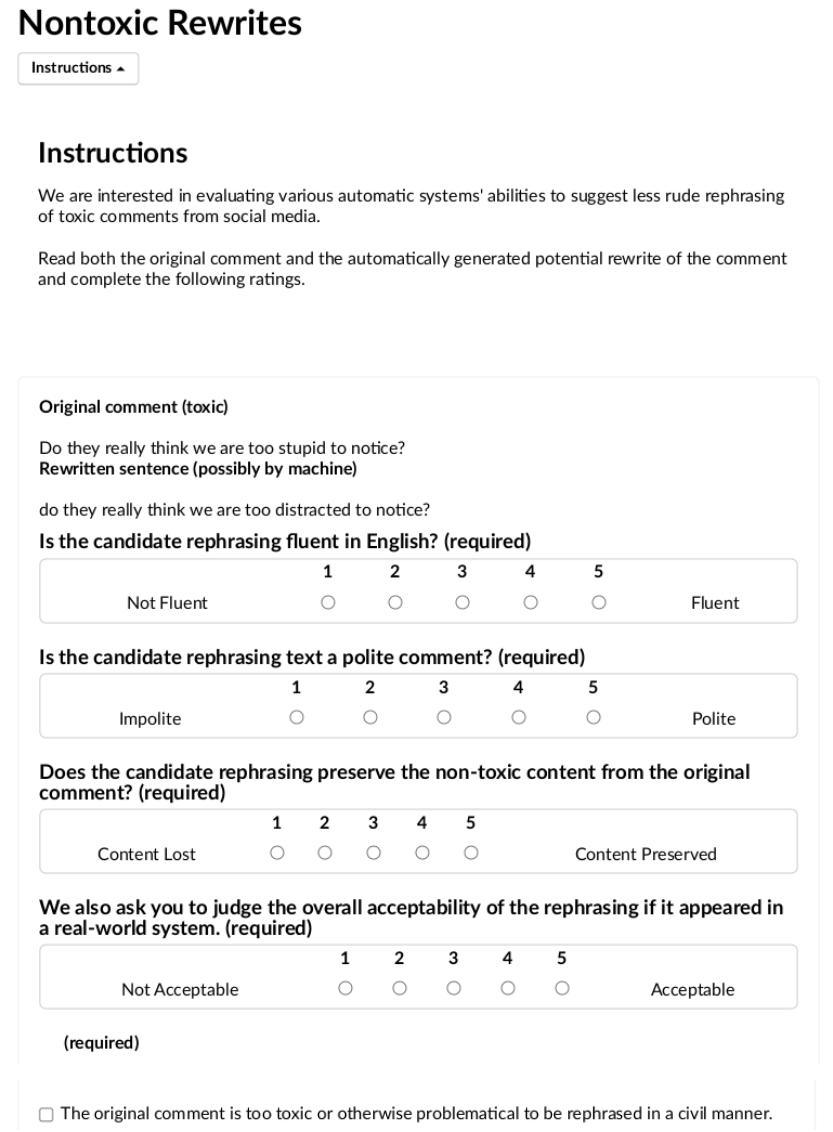}
\vspace*{-0mm}
\caption{\small Guidelines provided to human crowd-workers on Appen, when they were asked to rate automatic civil rephrasings of toxic comments.}
\vspace*{-0mm}
\label{fig:guidelines_appen_2}
\end{figure*}

\begin{figure*}
\includegraphics[width=\textwidth]{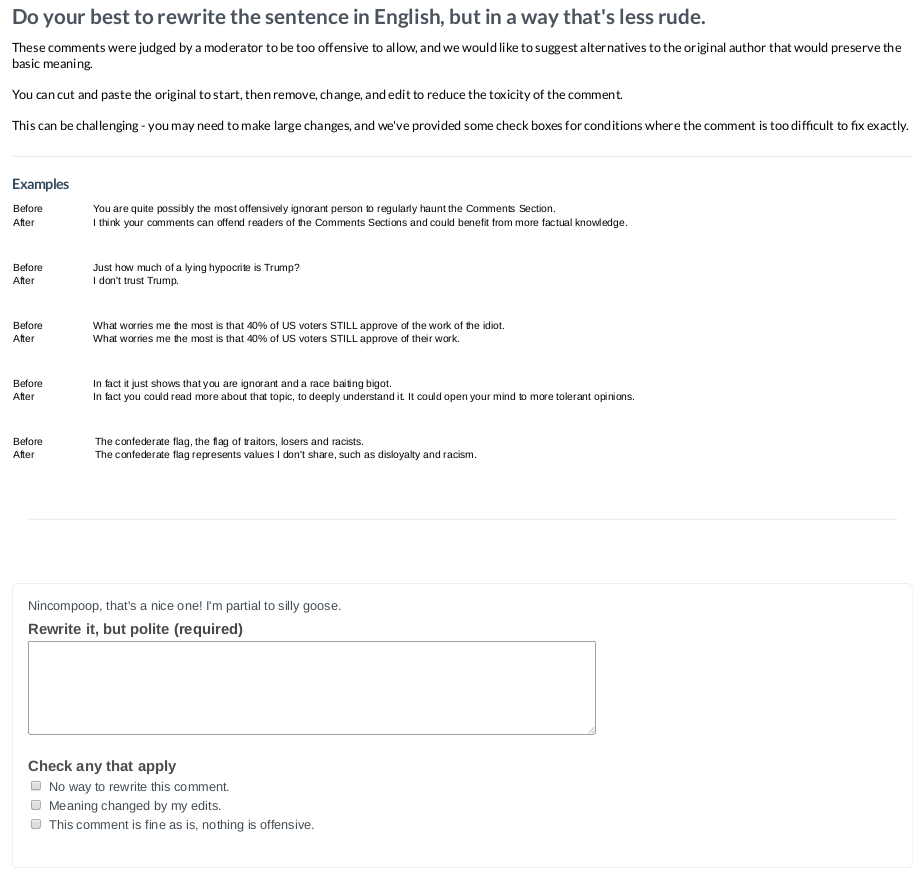}
\vspace*{-0mm}
\caption{\small Guidelines provided to human crowd-workers on Appen, when they were asked to rewrite toxic comments in a way that it is less rude.}
\vspace*{-0mm}
\label{fig:guidelines_appen}
\end{figure*}

Figure~\ref{fig:guidelines_appen_2} and Figure~\ref{fig:guidelines_appen} detail the guidelines we wrote on the crowdsourcing website Appen\footnote{\url{https://appen.com}}, when we asked human crowd-workers to rate automatic rephrasings and to rephrase toxic comments. Contributor level is set to level 3, which corresponds to the highest quality standard.

\end{document}